%% file: naaclhlt2019.tex
\documentclass[11pt,a4paper]{article}
\usepackage[hyperref]{naaclhlt2019}
\usepackage{times}
\usepackage{latexsym}
\usepackage{graphicx}
\usepackage[figuresleft]{rotating}

\usepackage{caption} 
\usepackage{subcaption}
\usepackage{listings}
\usepackage{color}
\usepackage{hyperref}
\usepackage{url}
\usepackage{todonotes}
\usepackage{ctable}
\usepackage{stfloats}

\usepackage{makecell}
\usepackage{tikz}
\usepackage{appendix}

\usepackage{tkz-graph}
\usetikzlibrary{fit}

\definecolor{codegreen}{rgb}{0,0.6,0}
\definecolor{codegray}{rgb}{0.5,0.5,0.5}
\definecolor{codepurple}{rgb}{0.58,0,0.82}
\definecolor{backcolour}{rgb}{0.97,0.97,0.97}
 
\lstdefinestyle{mystyle}{
    backgroundcolor=\color{backcolour},   
    commentstyle=\color{codegreen},
    keywordstyle=\color{orange},
    numberstyle=\tiny\color{codegray},
    stringstyle=\color{codepurple},
    basicstyle=\footnotesize,
    breakatwhitespace=false,         
    breaklines=true,                 
    captionpos=b,                    
    keepspaces=true,                 
    numbers=left,                    
    numbersep=5pt,                  
    showspaces=false,                
    showstringspaces=false,
    showtabs=false,                  
    tabsize=2
}
\usepackage{fdsymbol}

\newcommand{\cevb}[1]{\reflectbox{\scalebox{2}[1.5]{\ensuremath{\vec{\scalebox{0.5}[0.66]{\reflectbox{\ensuremath{#1}}}}}}}}
\newcommand{\vecb}[1]{\reflectbox{\reflectbox{\scalebox{2}[1.5]{\ensuremath{\vec{\scalebox{0.5}[0.66]{\ensuremath{#1}}}}}}}}

\aclfinalcopy 
\def\aclpaperid{1620} 

\newcommand\ourplans[0]{BestPlan}
\newcommand\ourbaseline[0]{StrongNeural}
\newcommand\ourrandom[1]{RandomPlan-#1}

\usepackage{fancyhdr}
\title{
    Step-by-Step: Separating Planning from Realization \\ 
    in Neural Data-to-Text Generation\thanks{\hspace{0.17cm}This research was supported in part by the German Research Foundation through the German-Israeli Project Cooperation (DIP, grant DA 1600/1-1) and by a grant from Theo Hoffenberg and Reverso.}
}

\author{Amit Moryossef$^\dagger$ \;\; Yoav Goldberg$^\dagger$$^\ddagger$ \;\; Ido Dagan$^\dagger$ \\
\texttt{amitmoryossef@gmail.com, \{yogo,dagan\}@cs.biu.ac.il} \\
\\ $^\dagger$Bar Ilan University, Ramat Gan, Israel \\
$^\ddagger$Allen Institute for Artificial Intelligence
}

\date{}

\begin{document}
\maketitle

\begin{abstract}
Data-to-text generation can be conceptually divided into two parts: ordering and structuring the information (planning), and generating fluent language describing the information (realization). Modern neural generation systems conflate these two steps into a single end-to-end differentiable system. We propose to split the generation process into a symbolic text-planning stage that is faithful to the input, followed by a neural generation stage that focuses only on realization. 
For training a plan-to-text generator, we present a method for matching reference texts to their corresponding text plans. For inference time, we describe a method for selecting high-quality text plans for new inputs. We implement and evaluate our approach on the WebNLG benchmark. Our results demonstrate that decoupling text planning from neural realization indeed improves the system's reliability and adequacy while maintaining fluent output. We observe improvements both in BLEU scores and in manual evaluations. Another benefit of our approach is the ability to output diverse realizations of the same input, paving the way to explicit control over the generated text structure.
\end{abstract}

\thispagestyle{fancy}

\section{Introduction}
Consider the task of data to text generation, as exemplified in the WebNLG corpus \cite{colin2016webnlg}. The system is given a set of RDF triplets describing facts (entities and relations between them) and has to produce a fluent text that is faithful to the facts. An example of such triplets is:\\[0.5em]
\texttt{John, birthPlace, London}\\
\texttt{John, employer, IBM}\\[0.5em]
With a possible output:\\[0.5em]
\emph{\small 1. John, who was born in London, works for IBM}.\\[0.5em]
Other outputs are also possible:\\[0.5em]
\emph{\small 2. John, who works for IBM, was born in London.}\\
\emph{\small 3. London is the birthplace of John, who works for IBM.}\\
\emph{\small 4. IBM employs John, who was born in London.}\\[0.5em]
These variations result from different ways of structuring the information: choosing which fact to mention first, and in which direction to express each fact. Another choice is to split the text into two different sentences, e.g., \\[0.5em]
\emph{\small 5. John works for IBM. John was born in London.}\\[0.5em]
Overall, the choice of fact ordering, entity ordering, and sentence splits for these facts give rise to 12 different structures, each of them putting the focus on somewhat different aspect of the information. Realistic inputs include more than two facts, greatly increasing the number of possibilities.

Another axis of variation is in how to verbalize the information for a given structure. For example, (2) can also be verbalized as\\[0.5em] 
\emph{\small 2a. John works for IBM and was born in London.}\\[0.5em] and (5) as:\\[0.5em]
\emph{\small 5a. John is employed by IBM. He was born in London.}\\[0.5em]
We refer to the first set of choices (how to structure the information) as \emph{text planning} and to the second 
(how to verbalize a plan) as \emph{plan realization}.\footnote{Note that the variation from 5 to 5a includes the introduction of a pronoun. This is traditionally referred to as \emph{referring expression generation} (REG), and falls between the planning and realization stages. We do not treat REG in this work, but our approach allows natural integration REG systems' outputs.} 

The distinction between planning and realization is at the core of classic natural language generation (NLG) works \cite{reiter2000building,DBLP:journals/corr/GattK17}. However, a recent wave of \emph{neural NLG systems} ignores this distinction and treat the problem as a single end-to-end task of learning to map facts from the input to the output text \cite{gardent2017webnlg,duvsek2018findings}. These neural systems encode the input facts into an intermediary vector-based representation, which is then decoded into text. While not stated in these terms, the neural system designers hope for the network to take care of both the planning and realization aspect of text generation. A notable exception is the work of \citet{puduppully2018data}, who introduce a neural content-planning module in the end-to-end architecture.

While the neural methods achieve impressive levels of output fluency, they also struggle to maintain coherency on longer texts \cite{wiseman2017challenges}, struggle to produce a coherent order of facts, and are often not faithful to the input facts, either omitting, repeating, hallucinating or changing facts (the NLG community refers to such errors as errors in \emph{adequacy} or \emph{correctness} of the generated text).
When compared to template-based methods, the neural systems win in fluency but fall short regarding content selection and faithfulness to the input \cite{puzikov2018e2e}. Also, they do not allow control over the output's structure.
We speculate that this is due to demanding too much of the network: while the neural system excels at capturing the language details required for fluent realization, they are less well equipped to deal with the higher levels text structuring in a consistent and verifiable manner.

\paragraph{Proposal} we propose an explicit, symbolic, text planning stage, whose output is fed into a neural generation system. The text planner determines the information structure and expresses it unambiguously---in our case as a sequence of ordered trees. This stage is performed symbolically and is guaranteed to remain faithful and complete with regards to the input facts. Once the plan is determined,\footnote{The exact plan can be determined based on a data-driven scoring function that ranks possible suggestions, as in this work, or by other user provided heuristics or a trained ML model. The plans' symbolic nature and precise relation to the input structures allow verification of their correctness.} a neural generation system is used to transform it into fluent, natural language text. By being able to follow the plan structure closely, the network is alleviated from the need to determine higher-level structural decisions and can track what was already covered more easily. This allows the network to perform the task it excels in, producing fluent, natural language outputs.

We demonstrate our approach on the WebNLG corpus and show it results in outputs which are as fluent as neural systems, but more faithful to the input facts. 
The method also allows explicit control of the output structure and the generation of diverse outputs (some diversity examples are available in the Appendix).
We release our code and the corpus extended with matching plans in \url{https://github.com/AmitMY/chimera}.

\section{Overview of the Approach}\label{sec:preliminaries}
\paragraph{Task Description}
Our method is concerned with the task of generating texts from inputs in the form of RDF sets. 
Each input can be considered as a graph, where the entities are nodes, and the RDF relations are directed labeled edges. 
Each input is paired with one or more reference texts describing these triplets. The reference can be either a single sentence or a sequence of sentences. 
Formally, each input $G$ consists of a set of triplets of the form $(s_i,r_i,o_i)$, where $s_i,o_i \in V$ (``subject'' and ``object'') correspond to entities from DBPedia, and $r_i \in R$ is a labeled DBPedia relation ($V$ and $R$ are the sets of entities and relations, respectively). 
For example, Figure \ref{fig:graph-original} shows a triplet set $G$ and Figure \ref{fig:graph-text} shows a reference text.
We consider the data set as a set of input-output pairs $(G,\text{ref})$, where the same $G$ may appear in several pairs, each time with a different reference.

\paragraph{Method Overview}
We split the generation process into two parts: text planning and sentence realization. Given an input $G$, we first generate a text plan $plan(G)$ specifying the division of facts to sentences, the order in which the facts are expressed in each sentence, and the ordering of the sentences. This data-to-plan step is non-neural (Section \ref{sec:plan-structure}).
Then, we generate each sentence according to the plan. This plan-to-sentence step is achieved through an NMT system (Section \ref{sec:plan-neural}). 

\input{inc/process.tex}
Figure \ref{fig:sentence-plan} demonstrates the entire process.

To facilitate our plan-based architecture, we devise a method to annotate $(G, ref)$ pairs with the corresponding plans (Section \ref{sec:plan-train}), and use it to construct a dataset which is used to train the plan-to-text translation. The same dataset is also used to devise a plan selection method (Section \ref{sec:plan-ranking}).

\paragraph{General Applicability}
It is worth considering the dataset-specific vs. general applicability aspects of  our method. On the low-level details, this work is very much dataset dependent. We show how to represent plans for specific datasets, and, importantly for this work, how to automatically construct plans for this dataset given inputs and expected natural language outputs. The method of plan construction will likely not generalize ``as is'' to other datasets, and the plan structure itself may also be found to be lacking for more demanding generation tasks. However, on a higher level, our proposal is very general: intermediary plan structures can be helpful, and one should consider ways of obtaining them, and of using them. In the short term, this will likely take the form of ad-hoc explorations of plan structures for specific tasks, as we do here, to establish their utility. In the longer term, research may evolve to looking into how general-purpose plan are structured. Our main message is that the separation of planning from realization, even in the context of neural generation, is a useful one to be considered.

\section{Text Planning} \label{sec:plan-structure}

\paragraph{Plan structure}
Our text plans capture the division of facts to sentences and the ordering of the sentences. Additionally, for each sentence, the plan captures
(1) the ordering of facts within the sentence; 
(2) The ordering of entities within a fact, which we call the \emph{direction} of the relation. For example, the \texttt{\{A, location, B\}} relation can be expressed as either \emph{A is located in B} or \emph{B is the location of A}; 
(3) the structure between facts that share an entity, namely chains and sibling structures as described below.

\input{inc/structure.tex}

A text plan is modeled as a sequence of sentence plans, to be realized in order. Each sentence plan is modeled as an ordered tree, specifying the structure in which the information should be realized. Structuring each sentence as a tree enables a clear succession between different facts through shared entities. Our text-plan design assumes that each entity is mentioned only once in a sentence, which holds in the WebNLG corpus.
The ordering of the entities and relations within a sentence is determined by a pre-order traversal of the tree.

Figure \ref{fig:sentence-plan}b shows an example of a text plan. 
Formally, given the input $G$, a text plan $T$ is a sequences of sentence plans $T=s_1,...,s_{N_T}$. A sentence plan $s$ is a labeled, ordered tree, with arcs of the form $(h,\ell,m)$, where $h,m \in V$ are head and modifier nodes, each corresponding to an input entity, and $\ell=(r,d)$ is the relation between nodes, where $r \in R$ is the RDF relation, and $d \in \{\rightarrow,\leftarrow\}$ denotes the direction in which the relation is expressed: $d=\rightarrow$ if $(h,r,m) \in G$, and $d=\leftarrow$ if $(m,r,h)\in G$.
A text plan $T$ is said to match an input $G$ iff every triplet $(s,r,o)$ in $G$ is expressed in $T$ exactly once, either as an edge $(s, (r,\rightarrow), o)$ or as an edge $(o, (r_i, \leftarrow), s)$.

Chains $(h,\ell_1,m), (m,\ell_2,x)$ represent a succession of facts that share a middle entity (Figure \ref{fig:structure-chain}), while siblings --- nodes with the same parent --- $(h,\ell_1,m_1),(h,\ell_2,m_2)$ represents a succession of facts about the same entity (Figure \ref{fig:structure-siblings}). Sibling and chain structures can be combined (Figure \ref{fig:structure-combined}).
An example of an input we addressed in the WebNLG corpus, and matching text plan is given in Figure \ref{fig:graph-plan}.\\
\textbf{Exhaustive generation } For small-ish input graphs $G$---such as those in the WebNLG task we consider here---it is trivial to generate all possible plans by first considering all the ways of grouping the input into sets, 
then from each set generating all possible trees by arranging it as an undirected graph and performing several DFS traversals starting from each node, where each DFS traversal follows a different order of children.\footnote{If a graph includes a cycle (0.4\% of the graphs in the WebNLG corpus contain cycles) we skip it, as it is guaranteed that a different split will result in cycle-free graphs.}

\subsection{Adding Plans to Training Data}\label{sec:plan-train}
While the input RDFs and references are present in the training dataset, the plans are not. We devise a method to recover the latent plans for most of the input-reference pairs in the training set, constructing a new dataset of $(G, ref, T)$ triplets of inputs, reference texts, and corresponding plans.

We define the reference $ref$, and the text-plan $T$ to be consistent with each other iff (a) they exhibit the same splitting into sentences---the facts in every sentence in $ref$ are grouped as a sentence plan in $T$, and (b) for each corresponding sentence and sentence-plan, the order of the entities is identical.

The matching of plans to references is based on the observations that (a) it is relatively easy to identify entities in the reference texts, and a pair of entities in an input is unique to a fact; (b) it is relatively easy to identify sentence splits; (c) a reference text and its matching plan must share the same entities in the same order, and with the same sentence splits.\\
\textbf{Sentence split consistency }
We define a set of triplets to be \emph{potentially consistent} with a sentence
iff each triplet contains at least one entity from the sentence (either its subject or object appear in the sentence), and each entity in the sentence is covered by at least one triplet.
Given a reference text, we split it into sentences using NLTK \cite{bird2004nltk}, and look for divisions of $G$ into disjoint sets such that each set is consistent with a corresponding sentence. For each such division, we consider the exhaustive set of all induced plans.\\
\textbf{Facts order consistency } 
A natural criterion would be to consider a reference sentence and a sentence-plan originating from the corresponding RDF as matching iff the sets of entities in the sentence and the plan are identical, and all entities appear in the  same order.\footnote{An additional constraint is that no two triplets in the RDFs set share the same entities. This is to ensure that if two entities appeared in a structure, only one relation could have been expressed there. This almost always holds in the WebNLG corpus, failing on only 15 out of 6,940 input sets.}
Based on this, we could represent each sentence and each plan as a sequence of entities, and verify the sequences match.

However, using this criterion is complicated by the fact that it is not trivial to map between the entities in the plan (that originate from the RDF triplets) and the entities in the text. In particular, due to language variability, the same plan entity may appear in several forms in the textual sentences. Some of these variations (i.e. ``A.F.C Fylde'' vs.  ``AFC Fylde'') can be recognized heuristically, while others require external knowledge (``UK conservative party'' vs. ``the Tories''), and some are ambiguous and require full-fledged co-reference resolution (``them'', ``he'', ``the former''). Hence, we relax our matching criterion to allow for possible unrecognized entities in the text.

Concretely, we represent each sentence plan as a sequence of its entities $(pe_1, ..., pe_k)$, and each sentence as the sequence of its entities which we managed to recognize and to match with an input entity $(se_1,...,se_m), m \leq k$.\footnote{We match plan entities to sentence entities using greedy string matching with Levenshtein distance   \cite{levenshtein1966binary} for each token and a manually tuned threshold for a match. 
While this approach results in occasional false positives, most cases are detected correctly. We match dates by using the 
\hyperlink{https://github.com/wanasit/chrono-python}{chrono-python} package that parses dates from natural language texts.}

We then consider a sentence and a sentence-plan to be \emph{consistent} if the following two conditions hold:
(1) The sentence entities $(se_1, ..., se_m)$ are a proper sub-sequence of the plan entities $(pe_1, ..., pe_k)$; and (2) each of the remaining entities in the plan already appeared previously in the plan. The second condition accounts for the fact that most un-identified entities are due to pronouns and similar non-lexicalized referring expressions, and that these only appear after a previous occurrence of the same entity in the text.%
\footnote{A sensible alternative would be to use a coreference resolution system at this stage. In our case it turned out to not help, and even performed somewhat worse.}

\subsection{Test-time Plan Selection}\label{sec:plan-ranking}

To select the plan to be realized, we propose a mechanism for ranking the possible plans. Our plan scoring method is a product-of-experts model, where each expert is a conditional probability estimate for some property of the plan. The conditional probabilities are MLE estimates based on the plans in the training set constructed in section \ref{sec:plan-train}. Estimates involving relation names are smoothed using Lidstone smoothing to account for unseen relations.
We use the following experts:

\noindent\textbf{Relation direction} 
For every relation $r\in R$, we compute its probability to be expressed in the plan in its original order ($d=\rightarrow)$ or in the reverse order ($d=\leftarrow$): $p_{dir}(d=\rightarrow | R)$.
This captures the tendency of certain relations to be realized in the reversed order to how they are defined in the knowledge base.
For example, in the WebNLG corpus the relation ``manager'' is expressed as a variation of ``is managed by'' instead of one of ``is the manager of'' in 68\% of its occurrences ($p_{dir}(d=\leftarrow | \text{manager})=0.68$).

\noindent\textbf{Global direction} 
We find that while the probability of each relation to be realized in a reversed order is usually below 0.5, still in most plans of longer texts there are one or two relations that appear in the reversed order. 
We capture this tendency using an expert that considers the conditional probability $p_{gd}(nr=n | \; |G|)$ of observing $n$ reversed edges in an input with $|G|$ triplets.

\noindent\textbf{Splitting tendencies}
For each input size, we keep track of the possible ways in which the set of facts can be split to subsets of particular sizes. 
That is, we keep track of probabilities such as $p_s(s=[3,2,2] \;|\; 7)$ of realizing an input of 7 RDF triplets as three sentences, each realizing the corresponding number of facts.

\noindent\textbf{Relation transitions}
We consider each sentence plan as a sequence of the relation types expressed in it $r_1,\ldots,r_k$ followed by an EOS symbol, and compute the markov transition probabilities over this sequence: $p_{trans}(r_1,r_2,\ldots,r_k,EOS) = \prod_{i=1,k} p_t(r_{i+1}|r_{i})$. The expert is the product of the transition probabilities of the individual sentence plans in the text plan.
This captures the tendencies of relations to follow each other and in particular, the tendencies of related relations such as birth-place and birth-date to group, allowing their aggregation in the generated text (\emph{John was born in London on Dec 12th, 1980}).

Each of the possible plans are then scored based on the product of the above quantities.\footnote{We note that for an input of $n$ triplets, there are $\mathcal{O}(2^{2n} + n*n!)$ possible plans,
making this method prohibitive for even moderately sized input graphs. However, it is sufficient for the WebNLG dataset in which $n\leq 7$. For larger graphs, better plan scoring and more efficient search algorithms should be devised. We leave this for future work.}

The scores work well for separating good from lousy text plans, 
and we observe a threshold above which most generated plans result in adequate texts. 
We demonstrate in Section \ref{sec:results} that realizing highly-ranked plans manages to obtain good automatic realization scores.
We note that the plan in Figure \ref{fig:graph-plan} is the one our ranking algorithm ranked first for the input in Figure  \ref{fig:graph-original}.

\input{inc/random-plans.tex}

\paragraph{Possible Alternatives} In addition to the single plan selection, the explicit planning stage opens up additional possibilities. Instead of choosing and realizing a single plan, we can realize \textbf{a diverse set} of high-scoring plans, or realizing a random high-scoring plan, resulting in a diverse and less templatic set of texts across runs. 
This relies on the combination of two factors: the ability of the scoring component to select plans that correspond to plausible human-authored texts, and the ability of the neural realizer to faithfully realize the plan into fluent text. While it is challenging to directly evaluate the plans adequacy, we later show an evaluation of the plan realization component. Figure \ref{fig:random-linearized} shows three random plans for the same graph and their realizations. Further examples of the diversity of generation are given in the appendix. 

The explicit and symbolic planning stage also allows for \textbf{user control} over the generated text, either by supplying constraints on the possible plans (e.g., number of sentences, entities to focus on, the order of entities/relations, or others) or by supplying complete plans. We leave these options for future work.

\section{Plan Realization}\label{sec:plan-neural} 

For plan realization, we use an off-the-shelf vanilla neural machine translation (NMT) system to translate plans to texts.
The explicit division to sentences in the text plan allows us to realize each sentence plan individually which allows the realizer to follow the plan structure within each (rather short) sentence, reducing the amount of information that the model needs to remember. As a result, we expect a significant reduction in over- and under-generation of facts, which are common when generating longer texts. Currently, this comes at the expense of not modeling discourse structure (i.e., referring expressions). This deficiency may be handled by integrating the discourse into the text plan, or as a post-processing step.\footnote{Minimally, each entity occurrence can keep track of the number of times it was already mentioned in the plan. Other alternatives include using a full-fledged referring expression generation system such as NeuralREG \cite{ferreira2018neuralreg}}. We leave this for future work.

To use text plans as inputs to the NMT, we linearize each sentence plan by performing a pre-order traversal of the tree, while indicating the tree structure with brackets (Figure \ref{fig:graph-linearized}). The directed relations $(r,d)$ are expressed as a sequence of two or more tokens, the first indicating the direction and the rest expressing the relation.\footnote{We map DBPedia relations to sequences of tokens by splitting on underscores and CamelCase.}  
Entities that are identified in the reference text are replaced with single, entity-unique tokens. This allows the NMT system to copy such entities from the input rather than generating them.
Figure \ref{fig:graph-text} is an example of possible text resulting from such linearization.

\paragraph{Training details} 
We use a standard NMT setup with a copy-attention mechanism \cite{gulcehre2016pointing}\footnote{Concretely, we use the OpenNMT toolkit \cite{klein2017opennmt} with the \texttt{copy\_attn} flag. Exact parameter values are detailed in the appendix.}
and the pre-trained GloVe.6B word embeddings\footnote{\url{nlp.stanford.edu/data/glove.6B.zip}} \cite{pennington2014glove}.  The pre-trained embeddings are used to initialize the relation tokens in the plans, as well as the tokens in the reference texts. 

\paragraph{Generation details} 
We translate each sentence plan individually. Once the text is generated, we replace the entity tokens with the full entity string as it appears in the input graph, and lexicalize all dates as \emph{Month DAY+ordinal, YEAR} (i.e., \emph{July 4th, 1776}) and for numbers with units (i.e., \emph{``5''(minutes)}) we remove the parenthesis and quotation marks (\emph{5 minutes}).

\section{Experimental Setup}
The WebNLG challenge \cite{colin2016webnlg} consists of mapping sets of RDF triplets to text including referring expression generation, aggregation, lexicalization, surface realization, and sentence segmentation. It contains sets with up to 7 triplets each along with one or more reference texts for each set. The test set is split into two parts: \emph{seen}, containing inputs created for entities and relations belonging to DBpedia categories that were seen in the training data, and \emph{unseen}, containing inputs extracted for entities and relations belonging to 5 unseen categories. While the unseen category is conceptually appealing, we view the seen category as the more relevant setup: generating fluent, adequate and diverse text for a mix of known relation types is enough of a challenge also without requiring the system to invent verbalizations for unknown relation types. Any realistic generation system could afford to provide at least a few verbalizations for each relation of interest. We thus focus our attention mostly on the seen case (though our system does also perform well on the unseen case).

Following Section \ref{sec:plan-train}, we manage to match a consistent plan for $76\%$ of the reference texts and use these plan-text pairs to train the plan realization NMT component. Overall, the WebNLG training set contains $18,102$ RDF-text pairs while our plan-enhanced corpus contains $13,828$ plan-text pairs.\footnote{Note that this only affects the training stage. At test time, we do not require gold plans, and evaluate on all sentences.}

\paragraph{Compared Systems}
We compare to the best submissions in the WebNLG challenge \cite{gardent2017webnlg}: Melbourne, an end-to-end system that scored best on all categories in the automatic evaluation, and UPF-FORGe \cite{mille2017forge}, a classic grammar-based NLG system that scored best 
in the human evaluation.

Additionally, we developed an end-to-end neural baseline which outperforms the WebNLG neural systems. 
It uses a set encoder, an LSTM \cite{hochreiter1997long} decoder with attention \cite{attention}, a copy-attention mechanism \cite{gulcehre2016pointing} and a neural checklist model \cite{kiddon2016globally}, as well as applying entity dropout. The entity-dropout and checklist component are the key differentiators from previous systems. 
We refer to this system as \textbf{\ourbaseline}.

\section{Experiments and Results}\label{sec:results}

\subsection{Automatic Metrics}
We begin by comparing our plan-based system (\textbf{\ourplans}) to the state-of-the-art using the common automatic metrics:
 BLEU \cite{papineni2002bleu}, Meteor \cite{banerjee2005meteor}, ROUGE$_L$ \cite{lin2004rouge} and CIDEr \cite{vedantam2015cider}, using the \hyperlink{https://github.com/Maluuba/nlg-eval}{nlg-eval}\footnote{\url{https://github.com/Maluuba/nlg-eval}} tool \cite{sharma2017nlgeval} on the entire test set and on each part separately (seen and unseen).

In the original challenge, the best performing system in automatic metric was based on end-to-end NMT (\textbf{Melbourne}). Both the \textbf{\ourbaseline} and \textbf{\ourplans} systems outperform all the WebNLG participating systems on all automatic metrics (Table \ref{table:scores-all}).
\textbf{\ourplans} is competitive with \textbf{\ourbaseline} in all metrics, with small differences either way per metric.\footnote{At least part of the stronger results for \textbf{StrongNeural} can be attributed to its ability to generate referring expressions, which we currently do not support.}

\input{inc/automatic/all.tex}

\subsection{Manual Evaluation}
Next, we turn to manually evaluate our system's performance regarding faithfulness to the input on the one hand and fluency on the other. 
We describe here the main points of the manual evaluation setup, with finer details in the appendix.

\input{inc/human/example.tex}

\paragraph{Faithfulness} 
As explained in Section \ref{sec:plan-structure}, the first benefit we expect of our plan-based architecture is to make the neural system’s task simpler, helping it to remain faithful to the semantics expressed in the plan which in turn is guaranteed to be faithful to the original RDF input (by faithfulness, we mean expressing all facts in the graph and only facts from the graph: not dropping, repeating or hallucinating facts).  We conduct a manual evaluation over the seen portion of the WebNLG human evaluated test set (139 input sets). We compare \textbf{\ourplans} and \textbf{\ourbaseline}.\footnote{We do not evaluate \textbf{UPF-FORGe} as it is a verifiable grammar-based system that is fully faithful by design.} For each output text, we manually mark which relations are expressed in it, which are omitted, and which relations exist with the wrong lexicalization. We also count the number of relations the system over generated, either repeating facts or inventing new facts.\footnote{This evaluation was conducted by the first author, on a set of shuffled examples from the \textbf{\ourplans} and \textbf{\ourbaseline} systems, without knowing which outputs belongs to which system. We further note that evaluating for faithfulness requires careful attention to detail (making it less suitable for crowd-workers), but has a precise task definition which does not involve subjective judgment, making it possible to annotate without annotator biases influencing the results. We release our judgments for this stage together with the code.}

Table \ref{table:expert} shows the results. \textbf{\ourplans} reduces all error types compared to \textbf{\ourbaseline}, by $85\%$, $56\%$ and $90\%$ respectively.
While on-par regarding automatic metrics, \textbf{\ourplans} substantially outperforms the new state-of-the-art end-to-end neural system in semantic faithfulness. 

For example, Figure \ref{fig:compare} compares the output of \textbf{\ourbaseline} (\ref{fig:compare-baseline}) and \textbf{\ourplans} (\ref{fig:compare-plan}) on the last input in the seen test set (\ref{fig:compare-baseline}).
While both systems chose three sentences split and aggregated details about birth in one sentence and details about the occupation in another, \textbf{\ourbaseline} also expressed the information in chronological order. However,  \textbf{\ourbaseline} failed to generate facts 3 and 5. \textbf{\ourplans} made a lexicalization mistake in the third sentence by expressing ``October'' before the actual date, which is probably caused by faulty entity matching for one of the references, and (by design) did not generate any referring expression, which we leave for future work.
\input{inc/human/semantics.tex}

\paragraph{Fluency} 
\input{inc/human/fluency.tex}
Next, we assess whether our systems succeed at maintaining the high-quality fluency of the neural systems.
We perform pairwise evaluation via Amazon Mechanical Turk wherein each task the worker is presented with an RDF set (both in a graph form, and textually), and two texts in random order, one from \textbf{\ourplans}, the other from a competing system. 
We compare \textbf{\ourplans} against a strong end-to-end neural system (\textbf{\ourbaseline}), a grammar-based system which is the state-of-the-art in human evaluation (\textbf{UPF-FORGe}), and the human-supplied WebNLG references (\textbf{Reference}).
The workers were presented with three possible answers: \textbf{\ourplans} text is better (scored as 1), the other text is better (scored as -1), and both texts are equally fluent (scored as 0).
Table \ref{table:mturk} shows the average worker score given to each pair divided by the number of texts compared. 
\textbf{\ourplans} performed on-par with \textbf{\ourbaseline}, and surpassed the previous state-of-the-art \textbf{UPF-FORGe}. It, however, scored worse than the reference texts, which is expected given that it does not produce referring expressions.
Our approach manages to keep the same fluency level typical to end-to-end neural systems, thanks to the NMT realization component.

\subsection{Plan Realization Consistency}
We test the extent to which the realizer generates texts that are consistent with the plans. For several subsets of ranked plans (best plan, top $1\%$, and top $10\%$) for the seen and unseen test sets separately, 
we realize up to 100 randomly selected text-plans per input.
We realize each sentence plan and evaluate using two criteria: (1) Do all entities from the plan appear in the realization; (2) Like the consistency we defined above, do all entities appear in the same order in the plan and the realization.

Table \ref{table:coverage} indicates that for decreasingly probable plans our realizer does worse in the first criterion. However, for both parts of the test set,
if the realizer managed to express all of the entities, it expressed them in the requested order, meaning the outputs are consistent with plans.
This opens up a potential for user control and diverse outputs, by choosing different plans for realization.

\input{inc/automatic/coverage.tex}

Finally, we verify that the realization of potentially diverse plans is not only consistent with each given plan but also preserves output quality.
For each input, we realize a random plan from the top $10\%$. We repeat this process three times with different random seeds to generate different outputs, and mark these systems as \textbf{\ourrandom{1/2/3}}. 
Table \ref{table:scores-all} shows that these random plans maintain decent quality on the automatic metrics, with a limited performance drop, and the automatic score is stable across random seeds.\footnote{While the scores for the different sets are very similar, the plans are very different from each other. See for examples the plans in Figure \ref{fig:random-linearized}.}  

\section{Related Work}\label{sec:related}
Text planning is a major component in classic NLG. For example, \citet{stent2004trainable} shows a method of producing coherent sentence plans by exhaustively generating as many as 20 sentence plan trees for each document plan, manually tagging them, and learning to rank them using the RankBoost algorithm \cite{schapire1999brief}. Our planning approach is similar, but we only have a set of ``good'' reference plans without internal ranks. While the sentence planning decides on the aggregation, one crucial decision left is sentence order. We currently determine order based on a splitting heuristic which relies on the number of facts in every sentence, not on the content. \citet{lapata2003probabilistic} devised a probabilistic model for sentence ordering which correlated well with human ordering. Our plan selection procedure is admittedly simple, and can be improved by integrating insights from previous text planning works \cite{barzilay2006aggregation, konstas2012unsupervised, konstas2013inducing}.

Many generation systems \cite{gardent2017webnlg, duvsek2018findings} are based on a black-box NMT component, with various pre-processing transformation of the inputs (such as delexicalization) and outputs to aid the generation process.

Generation from structured data often requires referring to a knowledge base \cite{mei2015talk, kiddon2016globally, wen2015semantically}.
This led to input-coverage tracking neural components such as the checklist model \cite{kiddon2016globally} and copy-mechanism \cite{gulcehre2016pointing}. Such methods are effective for ensuring coverage and reducing the number of over-generated facts and are in some ways orthogonal to our approach. While our explicit planning stage reduces the amount of over-generation, our realizer may be further improved by using a checklist model.

More complex tasks, like RotoWire \cite{wiseman2017challenges} require modeling also document-level planning. \citet{puduppully2018data} explored a method to explicitly model document planning using the attention mechanism.

The neural text generation community has also recently been interested in ``controllable'' text generation \cite{hu2017toward}, where various aspects of the text (often sentiment) are manipulated \cite{ficler2017controlling} or transferred \cite{shen2017style, zhao2018adversarially, li2018delete}. In contrast, like in \cite{wiseman2018learning}, here we focused on controlling either the content of a generation or the way it is expressed by manipulating the sentence plan used in realizing the generation.

\section{Conclusion}
We proposed adding an explicit symbolic planning component to a neural data-to-text NLG system, which eases the burden on the neural component concerning text structuring and fact tracking. Consequently, while the plan-based system performs on par with a strong end-to-end neural system regarding automatic evaluation metrics and human fluency evaluation, it substantially outperforms the end-to-end system regarding faithfulness to the input. Additionally, the planning stage allows explicit user-control and generating diverse sentences, to be pursued in future work.

\clearpage

\appendix
\appendixpage
\addappheadtotoc
\section{Diverse Outputs}

We demonstrate the ability of the model to produce diverse outputs by showing examples of generation from graphs with 4, 5 or 6 edges. For each graph, we show every $k$th plan, where $k$ is chosen so that our 25 examples cover the top 10\% of the plans, and order them by the scores assigned to them by the scoring model (the score is shown to the right of each plan, as well as the rank in the list). Higher scoring plans correspond to more natural plans, according to our model, but all of them are viable options. Then, for each plan we show the corresponding text generated by the NMT model. This provides a glimpse of: (1) the quality of the scoring model; (2) the diversity of the plans; (3) the naturalness of the generation.

For the plans, color boxes indicate entities, and gray boxes around them indicate bracketing. Vertical bars indicate sentence splits.
For the generated text, each entity is underlines with the color corresponding to its box.


\input{inc/diversity/size-4/latex.tex}
\clearpage
\input{inc/diversity/size-5/latex.tex}
\clearpage
\input{inc/diversity/size-6/latex.tex}
\clearpage

\section{Manual Evaluation Setup}\label{app:manual}
When performing pairwise system comparisons, we show the user, for each set of RDFs, the two texts produced by the compared systems in random order, along with the RDF triplets in textual and image forms as a reference. For consistency, both texts are normalized by lower-casing and splitting tokens on punctuation.
The same interface is used for turkers (for the fluency task) and local annotators (for the faithfulness task).

\subsection{Fluency Evaluation by Crowd}\label{app:manual:turk}
We evaluate on the RDF sets in the original WebNLG manual evaluation setup. The task is performed by mechanical-turk workers. The workers are presented with the question:

\textbf{``Which text reads more fluently?''} \\
which can be answered by either \emph{Text 1}, \emph{Text 2} or \emph{Both are equally good or bad}.

We paid $0.08\$$ per hit, employing three workers on each. For qualification, workers were required to have over 98\% hit approval rate, and over 1000 approved hits. 

\subsection{Faithfulness Evaluation by Expert}\label{app:manual:expert}
To obtain reliable fine-grained evaluation of semantic faithfulness, the first author annotated the system outputs of \textbf{\ourbaseline} and \textbf{\ourplans}.
    
For each text, we present all the RDF input triplets, and ask the annotator to choose for each triplet one of three options: (1) This triplet is \textit{expressed} in the text; (2) This triplet is not expressed in the text (\textit{ommitted}); (3) The text expresses a relation between the two entities that is different than the one specified for them in the RDF triplet (\textit{wrong lexicalization}). Also, for each text, we ask the annotator to count the number of facts that were wrongly \textit{over generated}, counting both repeated facts and hallucinated ones.

\section{Training Parameters}
For the realization model we use the OpenNMT toolkit \cite{klein2017opennmt} with pre-trained GloVe.6B word embeddings \cite{pennington2014glove}, downloaded from \url{http://nlp.stanford.edu/data/glove.6B.zip}. We used the default parameters (except for the \texttt{-copy\_attn} flag). This corresponds to the following values:
\begin{itemize}
    \item train\_steps = 40000
    \item save\_checkpoint\_steps = 2000
    \item batch\_size = 16
    \item word\_vec\_size = 300
    \item layers = 3
    \item copy\_attn
    \item position\_encoding
\end{itemize}

\clearpage

\bibliography{naaclhlt2019}
\bibliographystyle{acl_natbib}

\end{document}

%% file: inc/process.tex
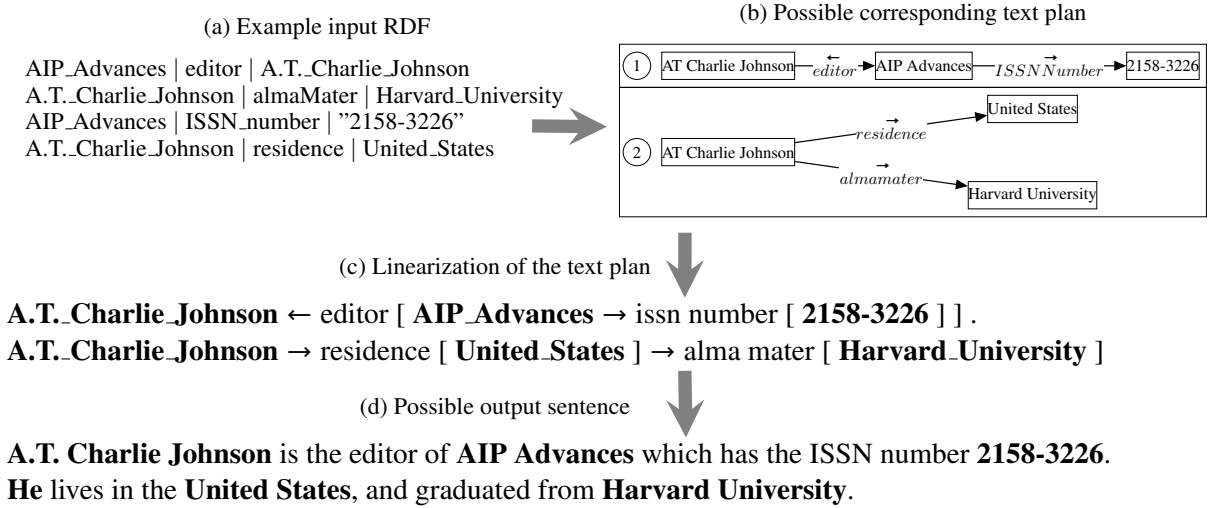
\begin{figure*}[t]
    \centering
    \begin{subfigure}{.49\textwidth}
        \caption{Example input RDF}

        \tikz[remember picture] \node[inner sep=0] (a) {
            \begin{footnotesize}
            \makecell[l]{
                AIP\_Advances \textbar\ editor \textbar\ A.T.\_Charlie\_Johnson\\
                A.T.\_Charlie\_Johnson \textbar\ almaMater \textbar\ Harvard\_University\\
                AIP\_Advances \textbar\ ISSN\_number \textbar\ "2158-3226"\\
                A.T.\_Charlie\_Johnson \textbar\ residence \textbar\ United\_States
            }
            \end{footnotesize}
        };
        \vspace{1cm}
        \label{fig:graph-original}
    \end{subfigure}%
    \begin{subfigure}{.49\textwidth}
        \centering
        \caption{Possible corresponding text plan}
        \tikz[remember picture] \node[inner sep=0] (b) {
        \resizebox{\linewidth}{!}{
            \begin{tikzpicture}
              \GraphInit[vstyle=Normal]
              \SetGraphUnit{1.5}
              \tikzset{VertexStyle/.append style={rectangle}}
            
              \Vertex[x=0,y=3,L=AT Charlie Johnson]{AT1}
              \Vertex[x=4.5,y=3]{AIP Advances}
              \Vertex[x=10,y=3]{2158-3226}
              
              \tikzset{VertexStyle/.append style={circle}}
              \Vertex[x=-2.1,y=3]{1}
              
              \Edge[style={->,>=triangle 45},label={$\protect\cevb{editor}$}](AT1)(AIP Advances)
              \Edge[style={->,>=triangle 45},label=$\protect\vecb{ISSN Number}$](AIP Advances)(2158-3226)
              \draw[draw=black] (-2.5,2.5) rectangle ++(13.5,1);
            
              \tikzset{VertexStyle/.append style={rectangle}}
            
              \Vertex[x=0,y=1,L=AT Charlie Johnson]{AT2}
              \Vertex[x=7,y=2]{United States}
              \Vertex[x=7,y=0]{Harvard University}
              
              \tikzset{VertexStyle/.append style={circle}}
              \Vertex[x=-2.1,y=1,style={circle}]{2}
            
              \Edge[style={->,>=triangle 45},label=$\protect\vecb{residence}$](AT2)(United States)
              \Edge[style={->,>=triangle 45},label=$\protect\vecb{alma mater}$](AT2)(Harvard University)
              \draw[draw=black] (-2.5,-0.5) rectangle ++(13.5,3);
            \end{tikzpicture}
        }
        };
        \label{fig:graph-plan}
    \end{subfigure}
    \begin{subfigure}{\textwidth}
        \caption{Linearization of the text plan\hspace{3cm}\phantom1}
        \tikz[remember picture] \node[inner sep=0] (c) {
            \makecell[l]{
                \textbf{A.T.\_Charlie\_Johnson} $\leftarrow$ editor [ \textbf{AIP\_Advances} $\rightarrow$ issn number [ \textbf{2158-3226} ] ] . \\
                \textbf{A.T.\_Charlie\_Johnson} $\rightarrow$ residence [ \textbf{United\_States} ] $\rightarrow$ alma mater [ \textbf{Harvard\_University} ]
            }
        };
        \label{fig:graph-linearized}
    \end{subfigure}
    \begin{subfigure}{\textwidth}
        \vspace{0.3cm}
        \caption{Possible output sentence\hspace{3cm}\phantom1}
        \tikz[remember picture] \node[inner sep=0] (d) {
            \makecell[l]{
                \textbf{A.T. Charlie Johnson} is the editor of \textbf{AIP Advances} which has the ISSN number \textbf{2158-3226}.\\
                \textbf{He} lives in the \textbf{United States}, and graduated from \textbf{Harvard University}.
            }
        };
        \label{fig:graph-text}
    \end{subfigure}
    
    \caption{Summary of our proposed generation process: the planner takes the input RDF triplets in (a), and generates the explicit plan in (b). The plan is then linearized (c) and passed to a neural generation system, producing the output (d).}
    \label{fig:sentence-plan}
    
    \begin{tikzpicture}[overlay, remember picture]
        \draw[-stealth, line width=5pt, gray] ([shift={(-5mm, -2mm)}]a.east)--([shift={(-1mm, 0.5mm)}]b.west);
        \draw[-stealth, line width=5pt, gray] ([shift={(-30mm, -2mm)}]b.south)--([shift={(17mm, 0mm)}]c.north);
        \draw[-stealth, line width=5pt, gray] ([shift={(17mm, 0mm)}]c.south)--([shift={(16.7mm, 0mm)}]d.north);
    \end{tikzpicture}
\end{figure*}

%% file: inc/structure.tex
\begin{figure}[ht]
    \begin{subfigure}{\linewidth}
        \resizebox{\linewidth}{!}{
        \begin{tikzpicture}
          \GraphInit[vstyle=Normal]
          \SetGraphUnit{1.5}
          \tikzset{VertexStyle/.append style={rectangle}}
        
          \Vertex[x=0,y=0]{John}
          \Vertex[x=4.5,y=0]{London}
          \Vertex[x=9,y=0]{England}
          \Edge[style={->,>=triangle 45},label=$ \protect\vecb{residence}$](John)(London)
          \Edge[style={->,>=triangle 45},label=$ \protect\cevb{capital}$](London)(England)
        \end{tikzpicture}
        }
        \caption{Chain: John lives in London, the capital of England.}
        \label{fig:structure-chain}
    \end{subfigure}\\
    \begin{subfigure}{\linewidth}
        \resizebox{0.58\linewidth}{!}{
        \begin{tikzpicture}
          \GraphInit[vstyle=Normal]
          \SetGraphUnit{1.5}
          \tikzset{VertexStyle/.append style={rectangle}}
        
          \Vertex[x=0,y=1]{John}
          \Vertex[x=4.5,y=2]{London}
          \Vertex[x=4.5,y=0]{Bartender}
          \Edge[style={->,>=triangle 45},label=$ \protect\vecb{residence}$](John)(London)
          \Edge[style={->,>=triangle 45},label=$ \protect\vecb{occupation}$](John)(Bartender)
        \end{tikzpicture}
        }        
        \caption{Sibling: John lives in London and works as a bartender.}
        \label{fig:structure-siblings}
    \end{subfigure}\\
    \begin{subfigure}{\linewidth}
        \resizebox{\linewidth}{!}{
        \begin{tikzpicture}
          \GraphInit[vstyle=Normal]
          \SetGraphUnit{1.5}
          \tikzset{VertexStyle/.append style={rectangle}}
        
          \Vertex[x=0,y=1]{John}
          \Vertex[x=4.5,y=2]{London}
          \Vertex[x=4.5,y=0]{Bartender}
          \Vertex[x=9,y=2]{England}
          \Edge[style={->,>=triangle 45},label=$ \protect\vecb{residence}$](John)(London)
          \Edge[style={->,>=triangle 45},label=$ \protect\vecb{occupation}$](John)(Bartender)
          \Edge[style={->,>=triangle 45},label=$ \protect\cevb{capital}$](London)(England)
        \end{tikzpicture}
        }        
        \caption{Combination: John lives in London, the capital of England, and works as a bartender.}
        \label{fig:structure-combined}
    \end{subfigure}%
    
    \caption{Fact construction structure.}
    \label{fig:structure-plan}
\end{figure}
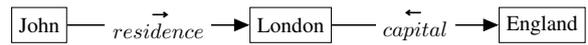
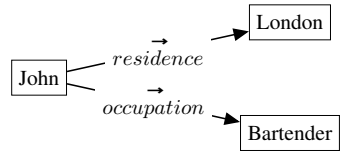
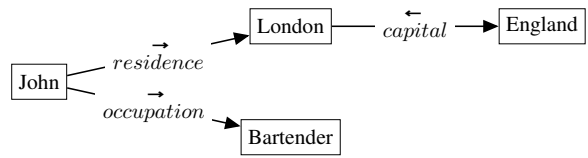

%% file: inc/random-plans.tex
\begin{figure*}[ht]
    \begin{subfigure}{\linewidth}
        (a) 
        \resizebox{\linewidth}{!}{
            \begin{minipage}{46em}
                \underline{\textbf{Dessert} $\leftarrow$ course [ \textbf{Bionico} $\rightarrow$ country [ \textbf{Mexico} ] $\rightarrow$ ingredient [ \textbf{Granola} ] $\rightarrow$ region [ \textbf{Jalisco} ] ] \hspace{2cm}}\\
                \emph{The \textbf{Dessert} \textbf{Bionico} requires \textbf{Granola} as one of its ingredients and originates from the \textbf{Jalisco} region of \textbf{Mexico} .}
            \end{minipage}
        }

    \end{subfigure}
    
    \vspace{1em}
    
    \begin{subfigure}{0.5\linewidth}
        (b)
        \resizebox{.95\linewidth}{!}{
            \begin{minipage}{23em}
                \textbf{Bionico} $\rightarrow$ country [ \textbf{Mexico} ] $\rightarrow$ region [ \textbf{Jalisco} ] . \\
                \underline{\textbf{Dessert} $\leftarrow$ course [ \textbf{Bionico} $\rightarrow$ ingredient [ \textbf{Granola} ] ]} 
                \emph{\textbf{Bionico} is a food found in the \textbf{Mexico} region \textbf{Jalisco}. \\
                The \textbf{Dessert} \textbf{Bionico} requires \textbf{Granola} as an ingredient.}
            \end{minipage}
        }
    \end{subfigure}
    \begin{subfigure}{0.5\linewidth}
        $\;$ (c)
        \resizebox{.95\linewidth}{!}{
            \begin{minipage}{23em}
                \textbf{Bionico} $\rightarrow$ ingredient [ \textbf{Granola} ] $\rightarrow$ course [ \textbf{Dessert} ] . \\
                \underline{\textbf{Bionico} $\rightarrow$ region [ \textbf{Jalisco} ] $\rightarrow$ country [ \textbf{Mexico} ]\hspace{.8cm}}
                \emph{\textbf{Bionico} contains \textbf{Granola} and is served as a \textbf{Dessert}.
                \textbf{Bionico} is a food found in the region of \textbf{Jalisco}, \textbf{Mexico}}
            \end{minipage}
        }
    \end{subfigure}
    
    \caption{Three random linearized plans for the same input graph, and their text realizations. All taken from the top 10\% scoring plans. (a) structures the output as a single sentence, while (b) and (c) as two sentences. The second sentence in (b) puts emphasis on Bionico being a dessert, while in (c) the emphasis is on the ingredients.}
    \label{fig:random-linearized}
\end{figure*}

%% file: inc/automatic/all.tex
\begin{table}[h]
        \centering
        \scalebox{0.7}{\begin{tabular}{|l|c|c|c|c|}
        \hline& \textbf{BLEU} & \textbf{METEOR} & \textbf{ROUGE\textsubscript{L}} & \textbf{CIDEr} \\ \hline
{\color{blue} UPF-FORGe$^\varheartsuit$} & 38.5 & 0.390 & 60.9 & 2.500\\ \hline
{\color{red} Melbourne$^\vardiamondsuit$} & 45.0 & 0.376 & 63.5 & 2.814\\ \specialrule{.1em}{.05em}{.05em} 
{\color{purple} \ourrandom{1}$^\spadesuit$} & 43.3 & 0.384 & 57.6 & 2.342\\ \hline
{\color{purple} \ourrandom{2}$^\spadesuit$} & 43.5 & 0.384 & 57.4 & 2.332\\ \hline
{\color{purple} \ourrandom{3}$^\spadesuit$} & 43.5 & 0.384 & 57.4 & 2.303\\ \hline
{\color{red} \ourbaseline$^\vardiamondsuit$} & 46.5 & \textbf{0.392} & \textbf{65.4} & \textbf{2.866}\\ \hline
{\color{purple} \ourplans$^\spadesuit$} & \textbf{47.4} & 0.391 & 63.1 & 2.692\\ \hline
\end{tabular}}
        \caption{Results for all categories. Team color indicates the type of system used 
        ({\color{red}NMT$^\vardiamondsuit$}, {\color{blue}Rule-Based$^\varheartsuit$}, {\color{purple}Rule-Based + NMT$^\spadesuit$}). 
        }
                \label{table:scores-all}

        \end{table}

%% file: inc/human/example.tex
\begin{figure*}[t]
	\centering
	\begin{subfigure}{.9\textwidth}
    	\fbox{\begin{subfigure}{.5\textwidth}
    		\begin{footnotesize}
            1. William\_Anders \textbar\ dateOfRetirement \textbar\ "1969-09-01" \\
            2. William\_Anders \textbar\ was selected by NASA \textbar\ 1963 \\
            3. William\_Anders \textbar\ timeInSpace \textbar\ "8820.0"(minutes) \\
            4. William\_Anders \textbar\ birthDate \textbar\ "1933-10-17"
    		\end{footnotesize}
    	\end{subfigure}
    	\begin{subfigure}{.5\textwidth}
    		\begin{footnotesize}
            5. William\_Anders \textbar\ occupation \textbar\ Fighter\_pilot \\
            6. William\_Anders \textbar\ birthPlace \textbar\ British\_Hong\_Kong \\
            7. William\_Anders \textbar\ was a crew member of \textbar\ Apollo\_8
    		\end{footnotesize}
    	\end{subfigure}}
		\caption{The last RDF in the seen test-set}
		\label{fig:compare-rdf}
	\end{subfigure}
	\begin{subfigure}{.9\linewidth}
	    \vspace{0.3cm}
	    \fbox{\begin{subfigure}{\textwidth}
    		\begin{footnotesize}
            	\textbf{William Anders} was born on \textbf{October 17th, 1933} in \textbf{British Hong Kong}.\\
                \textbf{He} was selected by nasa in \textbf{1963} and became a crew member on the \textbf{Apollo 8} flight mission.\\
                \textbf{He} retired on \textbf{September 1st, 1969}.
    		\end{footnotesize}
    	\end{subfigure}}
		\caption{Output from \ourbaseline}
		\label{fig:compare-baseline}
	\end{subfigure}
	\begin{subfigure}{.9\linewidth}
	    \vspace{0.3cm}
		\fbox{\begin{subfigure}{\textwidth}
    		\begin{footnotesize}
            	\textbf{William Anders} was a \textbf{fighter pilot} who joined nasa in \textbf{1963} and served as a crew member of \textbf{Apollo 8}.\\
                \textbf{William Anders} retired on \textbf{September 1st, 1969} and spent \textbf{8820.0 minutes} in space.\\
                \textbf{William Anders} was born in \textbf{British Hong Kong} on october \textbf{October 17th, 1933}.
    		\end{footnotesize}
    	\end{subfigure}}
		\caption{Output from \ourplans}
		\label{fig:compare-plan}
	\end{subfigure}%
	\caption{Comparing end-to-end neural generation with our plan based system.}
	\label{fig:compare}
\end{figure*}

%% file: inc/human/semantics.tex
\begin{table}[h]
\resizebox{\linewidth}{!}
{
\begin{tabular}{|l|l|l|}
\hline
                     & \textbf{\ourplans} & \textbf{\ourbaseline} \\ \hline
Expressed            & 417                & 360                   \\ \specialrule{.1em}{.05em}{.05em} 
Omitted              & 6                  & 41                    \\ \hline
Wrong-lexicalization & 17                 & 39                    \\ \hline
Over-generation      & 3                  & 29                    \\ \hline
\end{tabular}
}
\caption{Semantic faithfulness of each system regarding 440 RDF triplets from 139 input sets in the seen part of the manually evaluated test set.}

\label{table:expert}
\end{table}

%% file: inc/human/fluency.tex
\begin{table}[t]
\resizebox{\linewidth}{!}
{
\begin{tabular}{|l|l|l|l|}
\hline
            & \textbf{\ourbaseline} & \textbf{Reference} & \textbf{UPF-FORGe} \\ \hline
\textbf{\ourplans}   & -0.6\%      & -5.4\% &  +5.1\%   \\ \hline
\end{tabular}
}
\caption{MTurk average worker score for \emph{\ourplans} compared to each system. 
It is a worse than the reference texts, on-par with the neural end-to-end system, and a better than the previous state-of-the-art.}
\label{table:mturk}
\end{table}

%% file: inc/automatic/coverage.tex
\begin{table}[!t]
\centering
\resizebox{\linewidth}{!}
{\begin{tabular}{|l|l|l|l|l|l|l|}
\hline
       & \multicolumn{2}{c|}{\textbf{Best Plan}} & \multicolumn{2}{c|}{\textbf{Top 1\% Plans}} & \multicolumn{2}{c|}{\textbf{Top 10\% Plans}}     \\ \hline
       & \textbf{Entities} & \textbf{Order} & \textbf{Entities} & \textbf{Order} & \textbf{Entities} & \textbf{Order} \\ \hline
Seen   & 98.9\%                  & 100\%                & 95.9\%                  & 99.9\%                & 93.6\%                  & 100\%                \\ \hline
Unseen & 66.7\%                  & 100\%                & 45.3\%                  & 100\%                & 41.3\%                  & 100\%                \\ \hline

\end{tabular}}
\centering

\caption{Surface realizer performance. \emph{Entities:} Percent of sentence plans that were realized with all the requested entities. \emph{Order}: of the sentences that were realized with all requested entities, percentage of realizations that followed the requested entity order.}
\label{table:coverage}
\end{table}




%% file: inc/diversity/size-4/latex.tex

\begin{figure*}[!hb]
    \centering
    \includegraphics[width=\linewidth]{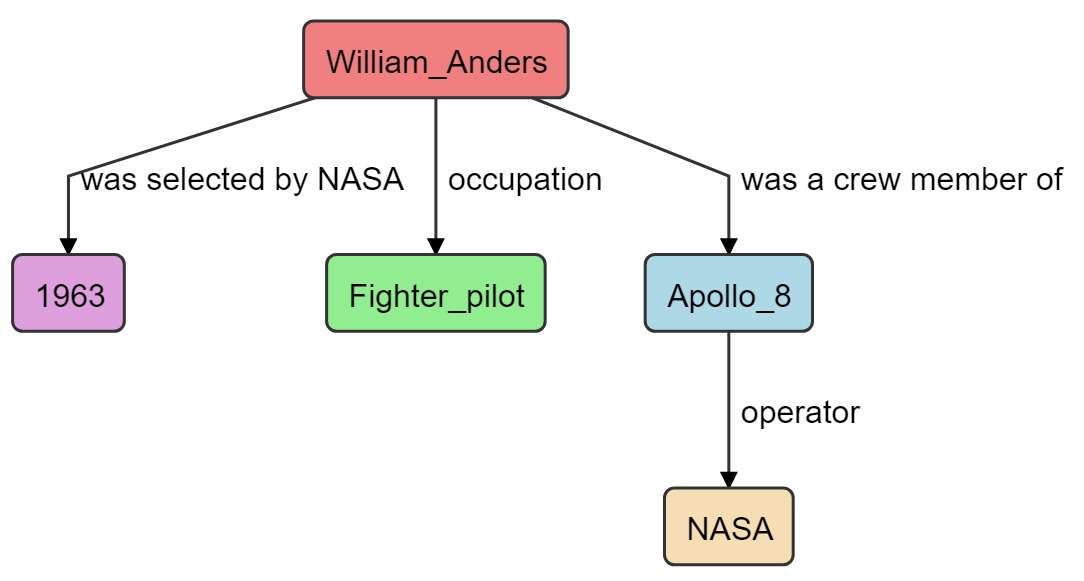}
    \caption{Example of a graph with 4 edges}
    \label{fig:diversity:size-4:graph}
\end{figure*}

\subsection{Example: Graph with 4 Edges}
\begin{sidewaysfigure*}
    \centering
    \includegraphics[width=\linewidth]{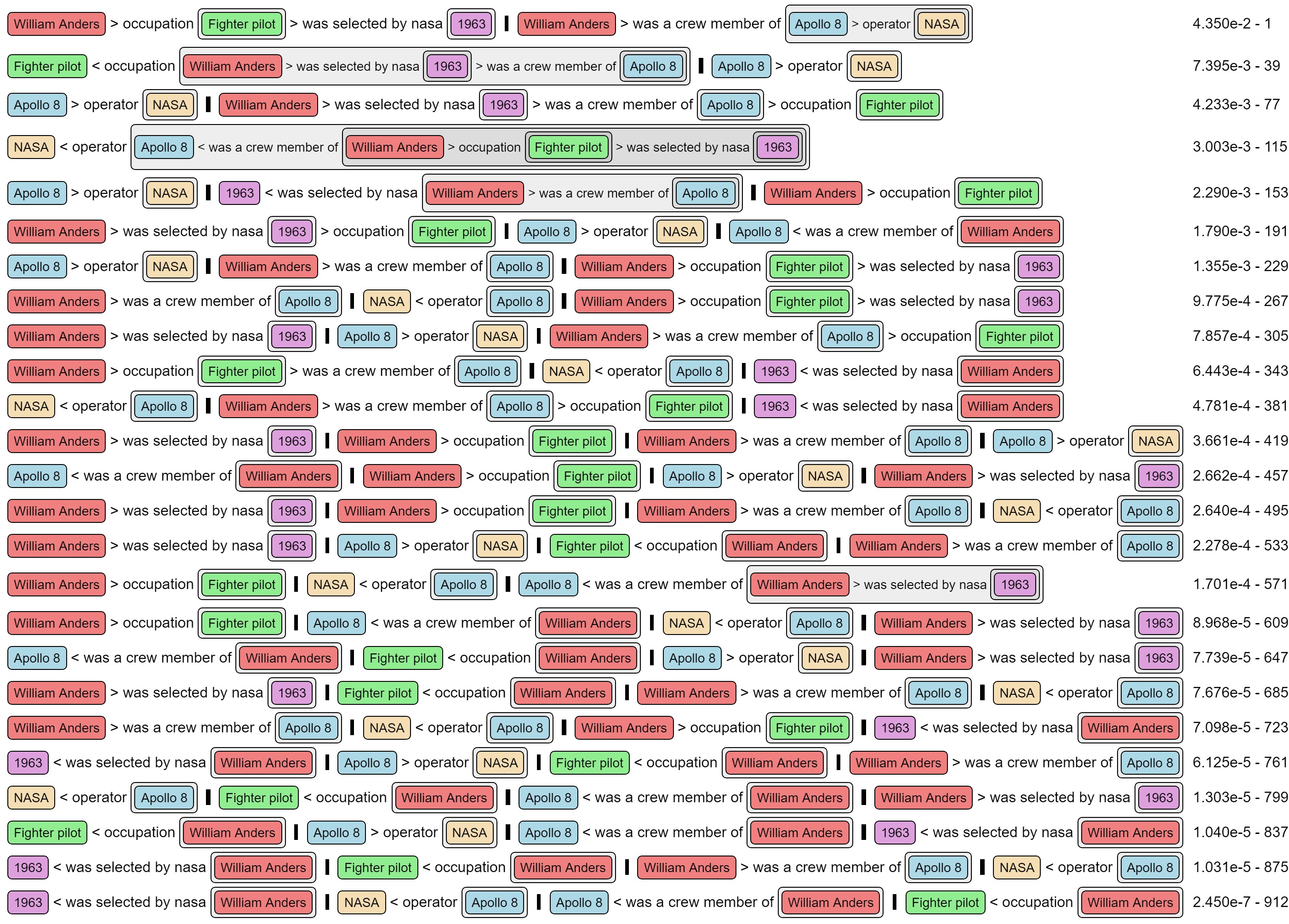}
    \caption{25 random linearized plans (out of 1,295 possible plans) for the graph in Figure \ref{fig:diversity:size-4:graph}, and their ranks and model scores.}
    \label{fig:diversity:size-4:plans}
\end{sidewaysfigure*}

\begin{sidewaysfigure*}
    \centering
    \includegraphics[width=0.95\linewidth]{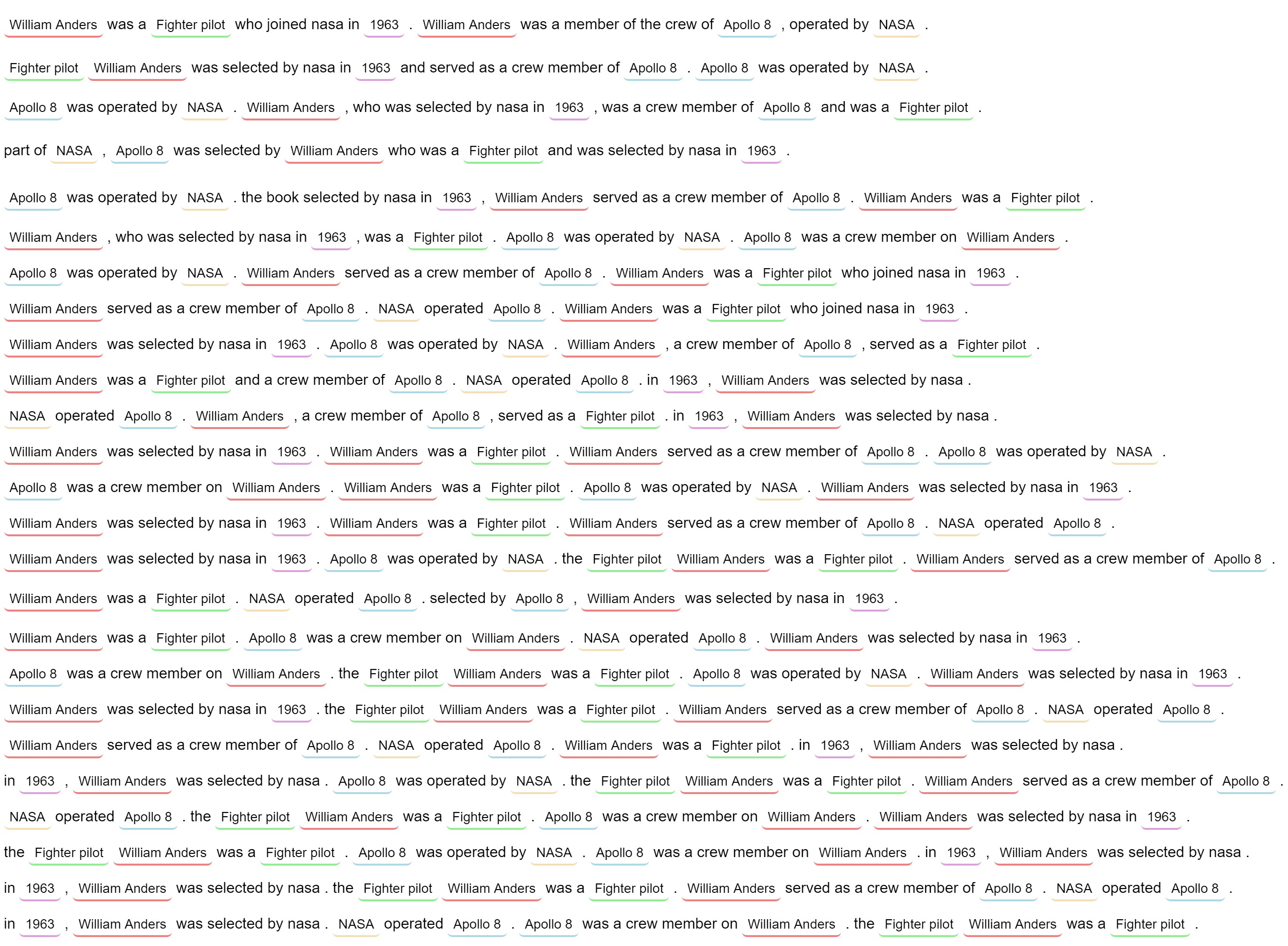}
    \caption{Realizations of the plans from Figure \ref{fig:diversity:size-4:plans} as produced by the NMT realizer.}
    \label{fig:diversity:size-4:output}
\end{sidewaysfigure*}

Figure \ref{fig:diversity:size-4:graph} shows a random 4-edge graph from the seen part of the test set. Figure \ref{fig:diversity:size-4:plans} shows the plans and Figure \ref{fig:diversity:size-4:output} the corresponding texts.



%% file: inc/diversity/size-5/latex.tex
\begin{figure*}[!hb]
    \centering
    \includegraphics[width=\linewidth]{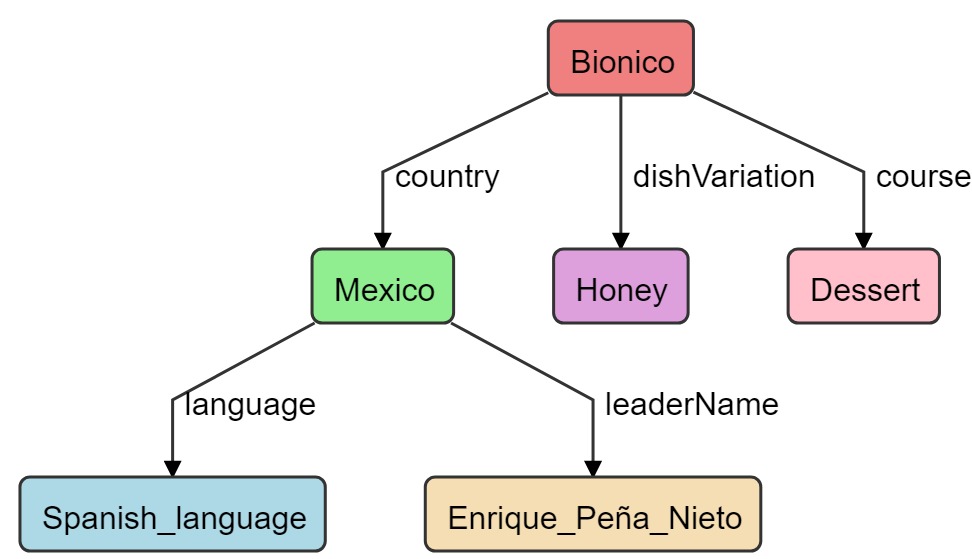}
    \caption{Example of a graph with 5 edges}
    \label{fig:diversity:size-5:graph}
\end{figure*}

\subsection{Example: Graph with 5 Edges}
\begin{sidewaysfigure*}
    \centering
    \includegraphics[width=0.87\linewidth]{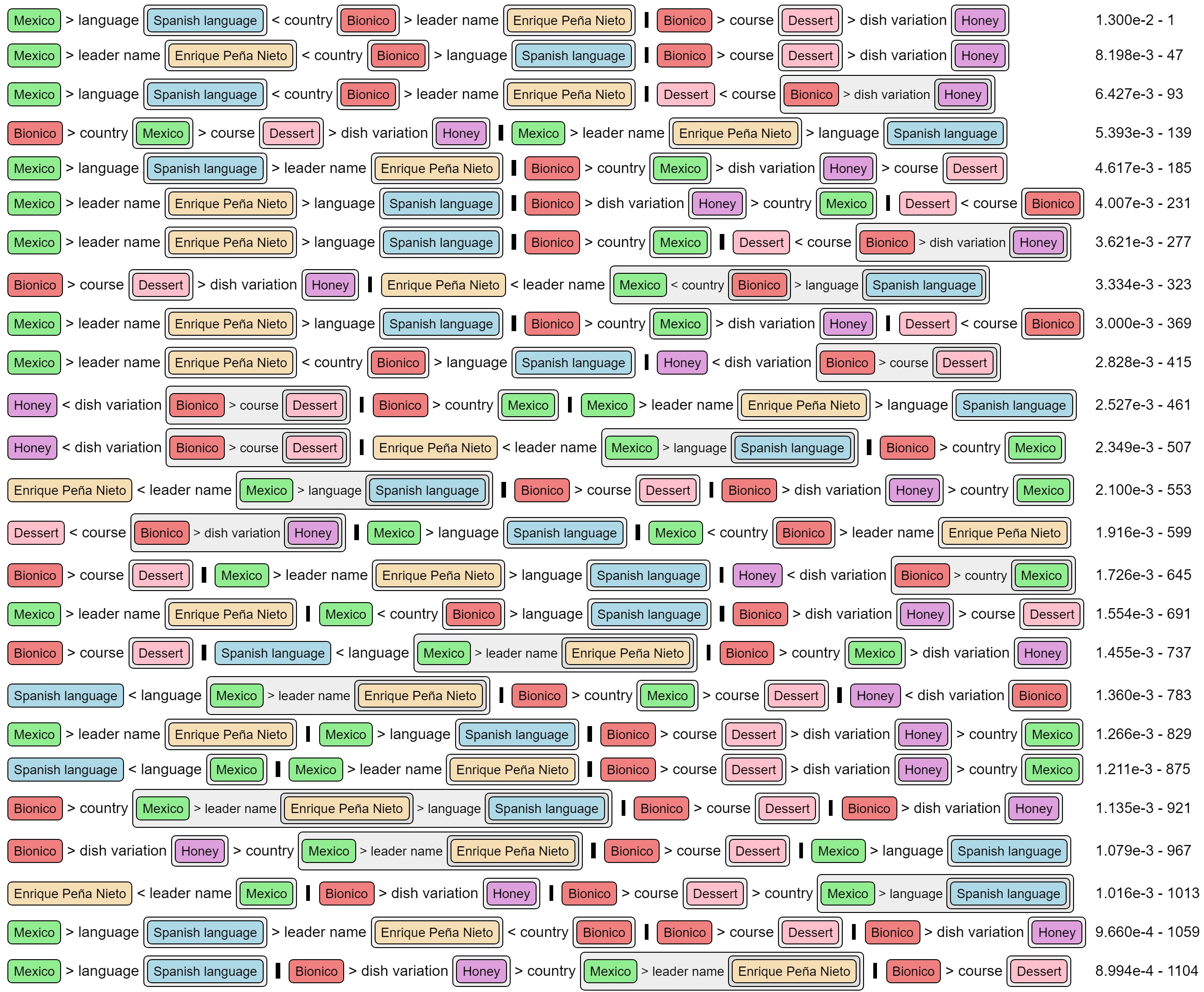}
    \caption{25 random linearized plans (out of 9,460 possible plans) for the graph in Figure \ref{fig:diversity:size-5:graph}, and their ranks and model scores.}
    \label{fig:diversity:size-5:plans}
\end{sidewaysfigure*}

\begin{sidewaysfigure*}
    \centering
    \includegraphics[width=\linewidth]{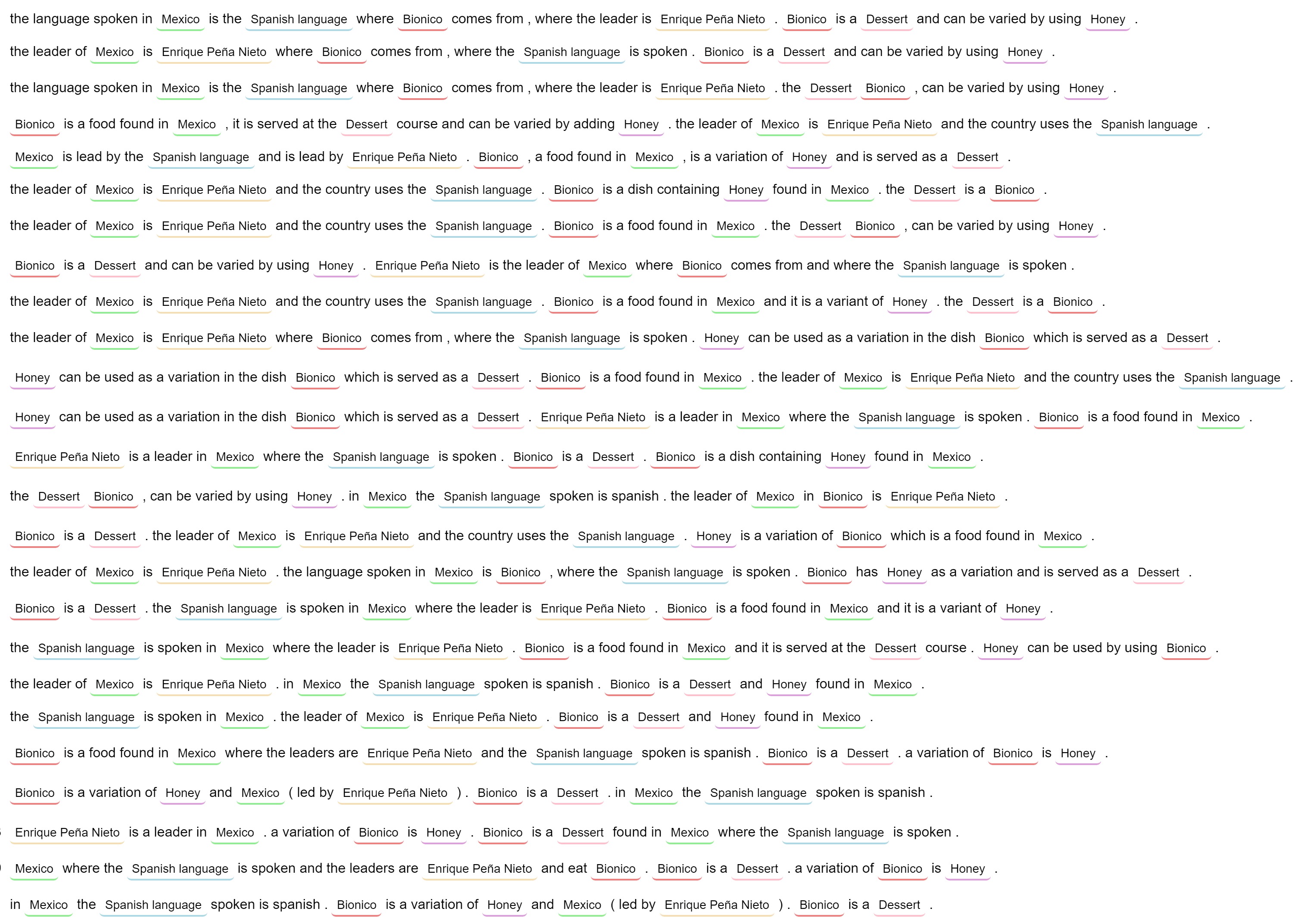}
    \caption{Realizations of the plans from Figure \ref{fig:diversity:size-5:plans} as produced by the NMT realizer.}
    \label{fig:diversity:size-5:output}
\end{sidewaysfigure*}

Figure \ref{fig:diversity:size-5:graph} shows a random 5-edge graph from the seen part of the test set. Figure \ref{fig:diversity:size-5:plans} shows the plans and Figure \ref{fig:diversity:size-5:output} the corresponding texts.




%% file: inc/diversity/size-6/latex.tex

\begin{figure*}[!hb]
    \centering
    \includegraphics[width=\linewidth]{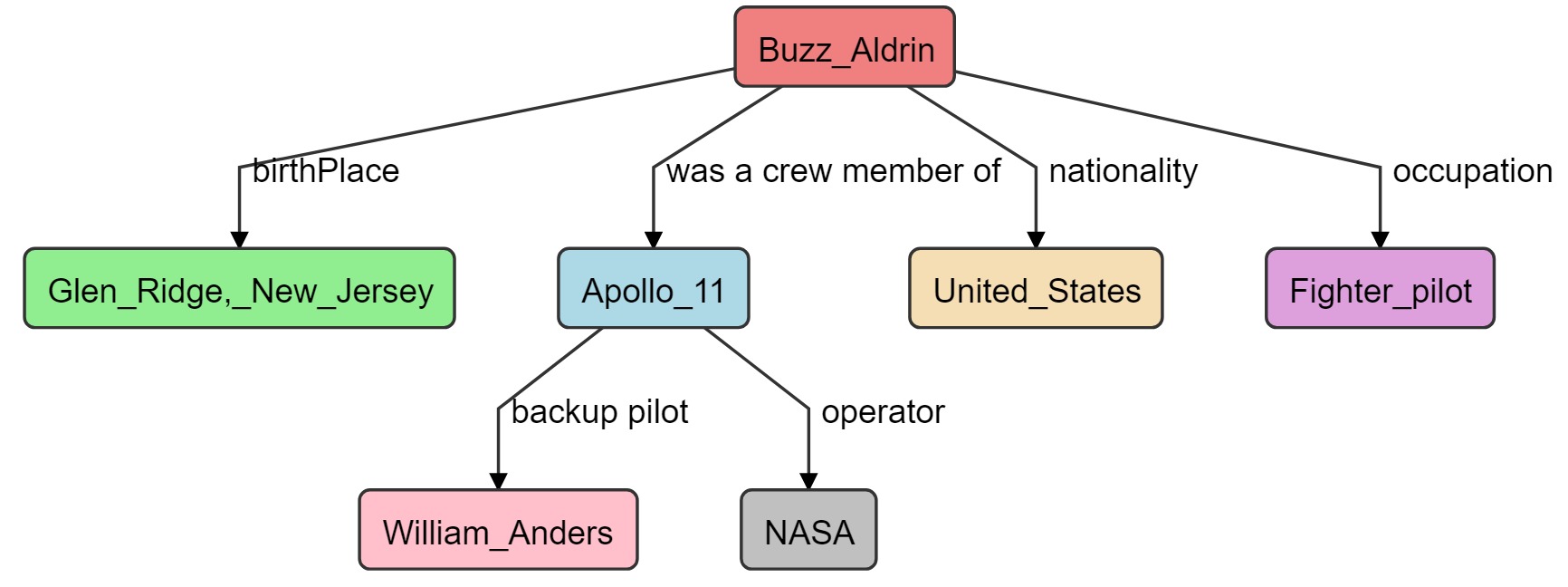}
    \caption{Example of a graph with 6 edges}
    \label{fig:diversity:size-6:graph}
\end{figure*}

\subsection{Example: Graph with 6 Edges}
\begin{sidewaysfigure*}
    \centering
    \includegraphics[width=\linewidth]{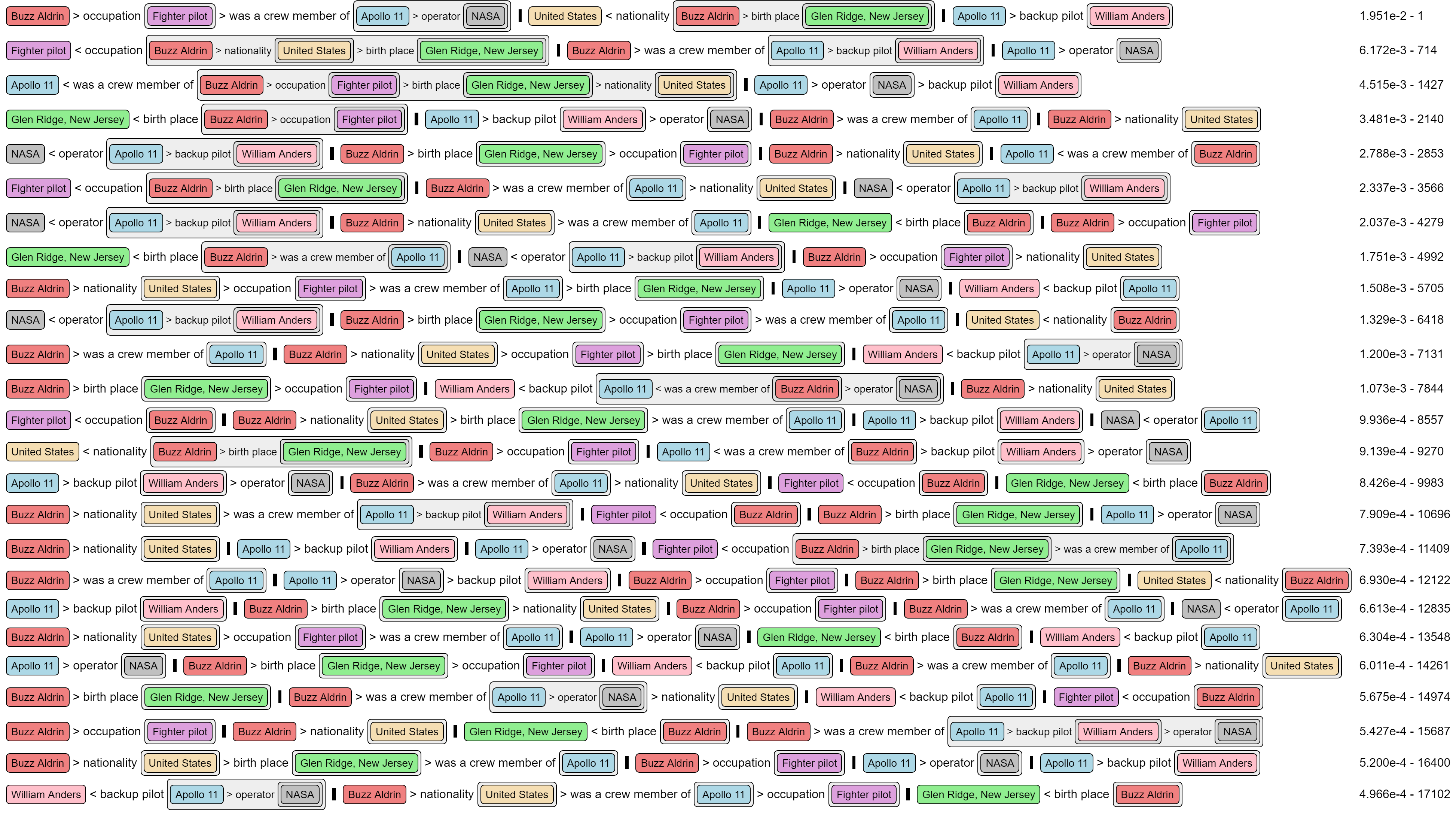}
    \caption{25 random linearized plans (out of 171,024 possible plans) for the graph in Figure \ref{fig:diversity:size-6:graph}, and their ranks and model scores.}
    \label{fig:diversity:size-6:plans}
\end{sidewaysfigure*}

\begin{sidewaysfigure*}
    \centering
    \includegraphics[width=\linewidth]{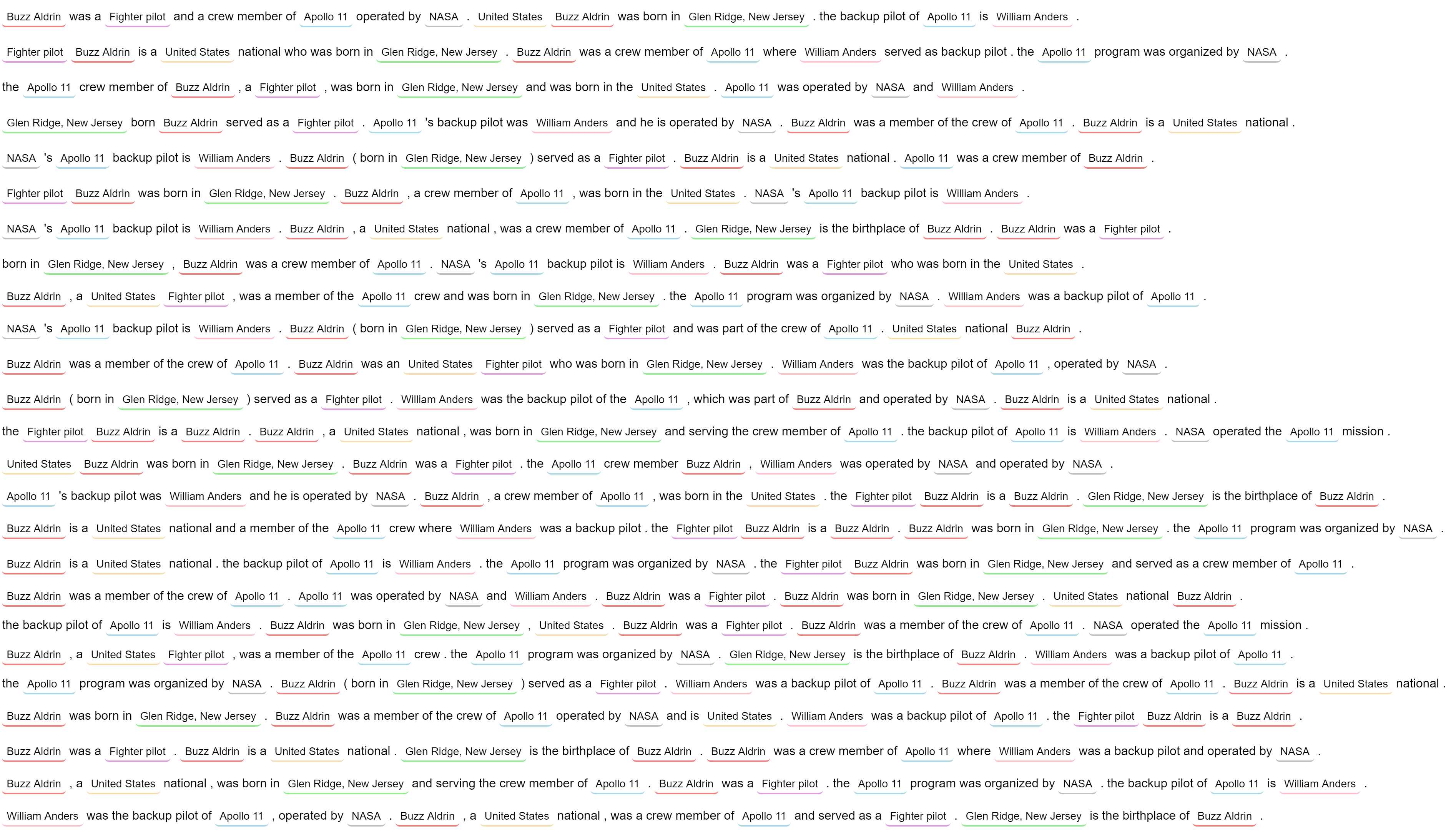}
    \caption{Realizations of the plans from Figure \ref{fig:diversity:size-6:plans} as produced by the NMT realizer.}
    \label{fig:diversity:size-6:output}
\end{sidewaysfigure*}

Figure \ref{fig:diversity:size-6:graph} shows a random 6-edge graph from the seen part of the test set. Figure \ref{fig:diversity:size-6:plans} shows the plans and Figure \ref{fig:diversity:size-6:output} the corresponding texts.



%% file: naaclhlt2019.bbl
\begin{thebibliography}{37}
\expandafter\ifx\csname natexlab\endcsname\relax\def\natexlab#1{#1}\fi

\bibitem[{Bahdanau et~al.(2014)Bahdanau, Cho, and Bengio}]{attention}
Dzmitry Bahdanau, Kyunghyun Cho, and Yoshua Bengio. 2014.
\newblock \href {http://arxiv.org/abs/1409.0473} {Neural machine translation by
  jointly learning to align and translate}.
\newblock \emph{CoRR}, abs/1409.0473.

\bibitem[{Banerjee and Lavie(2005)}]{banerjee2005meteor}
Satanjeev Banerjee and Alon Lavie. 2005.
\newblock Meteor: An automatic metric for mt evaluation with improved
  correlation with human judgments.
\newblock In \emph{Proceedings of the acl workshop on intrinsic and extrinsic
  evaluation measures for machine translation and/or summarization}, pages
  65--72.

\bibitem[{Barzilay and Lapata(2006)}]{barzilay2006aggregation}
Regina Barzilay and Mirella Lapata. 2006.
\newblock Aggregation via set partitioning for natural language generation.
\newblock In \emph{Proceedings of the main conference on Human Language
  Technology Conference of the North American Chapter of the Association of
  Computational Linguistics}, pages 359--366. Association for Computational
  Linguistics.

\bibitem[{Bird and Loper(2004)}]{bird2004nltk}
Steven Bird and Edward Loper. 2004.
\newblock Nltk: the natural language toolkit.
\newblock In \emph{Proceedings of the ACL 2004 on Interactive poster and
  demonstration sessions}, page~31. Association for Computational Linguistics.

\bibitem[{Colin et~al.(2016)Colin, Gardent, Mrabet, Narayan, and
  Perez-Beltrachini}]{colin2016webnlg}
Emilie Colin, Claire Gardent, Yassine Mrabet, Shashi Narayan, and Laura
  Perez-Beltrachini. 2016.
\newblock The webnlg challenge: Generating text from dbpedia data.
\newblock In \emph{Proceedings of the 9th International Natural Language
  Generation conference}, pages 163--167.

\bibitem[{Du{\v{s}}ek et~al.(2018)Du{\v{s}}ek, Novikova, and
  Rieser}]{duvsek2018findings}
Ond{\v{r}}ej Du{\v{s}}ek, Jekaterina Novikova, and Verena Rieser. 2018.
\newblock Findings of the e2e nlg challenge.
\newblock \emph{arXiv preprint arXiv:1810.01170}.

\bibitem[{Ferreira et~al.(2018)Ferreira, Moussallem, K{\'a}d{\'a}r, Wubben, and
  Krahmer}]{ferreira2018neuralreg}
Thiago~Castro Ferreira, Diego Moussallem, {\'A}kos K{\'a}d{\'a}r, Sander
  Wubben, and Emiel Krahmer. 2018.
\newblock Neuralreg: An end-to-end approach to referring expression generation.
\newblock \emph{arXiv preprint arXiv:1805.08093}.

\bibitem[{Ficler and Goldberg(2017)}]{ficler2017controlling}
Jessica Ficler and Yoav Goldberg. 2017.
\newblock Controlling linguistic style aspects in neural language generation.
\newblock \emph{arXiv preprint arXiv:1707.02633}.

\bibitem[{Gardent et~al.(2017)Gardent, Shimorina, Narayan, and
  Perez-Beltrachini}]{gardent2017webnlg}
Claire Gardent, Anastasia Shimorina, Shashi Narayan, and Laura
  Perez-Beltrachini. 2017.
\newblock The webnlg challenge: Generating text from rdf data.
\newblock In \emph{Proceedings of the 10th International Conference on Natural
  Language Generation}, pages 124--133.

\bibitem[{Gatt and Krahmer(2017)}]{DBLP:journals/corr/GattK17}
Albert Gatt and Emiel Krahmer. 2017.
\newblock \href {http://arxiv.org/abs/1703.09902} {Survey of the state of the
  art in natural language generation: Core tasks, applications and evaluation}.
\newblock \emph{CoRR}, abs/1703.09902.

\bibitem[{Gulcehre et~al.(2016)Gulcehre, Ahn, Nallapati, Zhou, and
  Bengio}]{gulcehre2016pointing}
Caglar Gulcehre, Sungjin Ahn, Ramesh Nallapati, Bowen Zhou, and Yoshua Bengio.
  2016.
\newblock Pointing the unknown words.
\newblock \emph{arXiv preprint arXiv:1603.08148}.

\bibitem[{Hochreiter and Schmidhuber(1997)}]{hochreiter1997long}
Sepp Hochreiter and J{\"u}rgen Schmidhuber. 1997.
\newblock Long short-term memory.
\newblock \emph{Neural computation}, 9(8):1735--1780.

\bibitem[{Hu et~al.(2017)Hu, Yang, Liang, Salakhutdinov, and
  Xing}]{hu2017toward}
Zhiting Hu, Zichao Yang, Xiaodan Liang, Ruslan Salakhutdinov, and Eric~P Xing.
  2017.
\newblock Toward controlled generation of text.
\newblock \emph{arXiv preprint arXiv:1703.00955}.

\bibitem[{Kiddon et~al.(2016)Kiddon, Zettlemoyer, and
  Choi}]{kiddon2016globally}
Chlo{\'e} Kiddon, Luke Zettlemoyer, and Yejin Choi. 2016.
\newblock Globally coherent text generation with neural checklist models.
\newblock In \emph{Proceedings of the 2016 Conference on Empirical Methods in
  Natural Language Processing}, pages 329--339.

\bibitem[{Klein et~al.(2017)Klein, Kim, Deng, Senellart, and
  Rush}]{klein2017opennmt}
Guillaume Klein, Yoon Kim, Yuntian Deng, Jean Senellart, and Alexander~M Rush.
  2017.
\newblock Opennmt: Open-source toolkit for neural machine translation.
\newblock \emph{arXiv preprint arXiv:1701.02810}.

\bibitem[{Konstas and Lapata(2012)}]{konstas2012unsupervised}
Ioannis Konstas and Mirella Lapata. 2012.
\newblock Unsupervised concept-to-text generation with hypergraphs.
\newblock In \emph{Proceedings of the 2012 Conference of the North American
  Chapter of the Association for Computational Linguistics: Human Language
  Technologies}, pages 752--761. Association for Computational Linguistics.

\bibitem[{Konstas and Lapata(2013)}]{konstas2013inducing}
Ioannis Konstas and Mirella Lapata. 2013.
\newblock Inducing document plans for concept-to-text generation.
\newblock In \emph{Proceedings of the 2013 Conference on Empirical Methods in
  Natural Language Processing}, pages 1503--1514.

\bibitem[{Lapata(2003)}]{lapata2003probabilistic}
Mirella Lapata. 2003.
\newblock Probabilistic text structuring: Experiments with sentence ordering.
\newblock In \emph{Proceedings of the 41st Annual Meeting on Association for
  Computational Linguistics-Volume 1}, pages 545--552. Association for
  Computational Linguistics.

\bibitem[{Levenshtein(1966)}]{levenshtein1966binary}
Vladimir~I Levenshtein. 1966.
\newblock Binary codes capable of correcting deletions, insertions, and
  reversals.
\newblock In \emph{Soviet physics doklady}, volume~10, pages 707--710.

\bibitem[{Li et~al.(2018)Li, Jia, He, and Liang}]{li2018delete}
Juncen Li, Robin Jia, He~He, and Percy Liang. 2018.
\newblock Delete, retrieve, generate: A simple approach to sentiment and style
  transfer.
\newblock \emph{arXiv preprint arXiv:1804.06437}.

\bibitem[{Lin(2004)}]{lin2004rouge}
Chin-Yew Lin. 2004.
\newblock Rouge: A package for automatic evaluation of summaries.
\newblock \emph{Text Summarization Branches Out}.

\bibitem[{Mei et~al.(2015)Mei, Bansal, and Walter}]{mei2015talk}
Hongyuan Mei, Mohit Bansal, and Matthew~R Walter. 2015.
\newblock What to talk about and how? selective generation using lstms with
  coarse-to-fine alignment.
\newblock \emph{arXiv preprint arXiv:1509.00838}.

\bibitem[{Mille et~al.(2017)Mille, Carlini, Burga, and Wanner}]{mille2017forge}
Simon Mille, Roberto Carlini, Alicia Burga, and Leo Wanner. 2017.
\newblock Forge at semeval-2017 task 9: Deep sentence generation based on a
  sequence of graph transducers.
\newblock In \emph{Proceedings of the 11th International Workshop on Semantic
  Evaluation (SemEval-2017)}, pages 920--923.

\bibitem[{Papineni et~al.(2002)Papineni, Roukos, Ward, and
  Zhu}]{papineni2002bleu}
Kishore Papineni, Salim Roukos, Todd Ward, and Wei-Jing Zhu. 2002.
\newblock Bleu: a method for automatic evaluation of machine translation.
\newblock In \emph{Proceedings of the 40th annual meeting on association for
  computational linguistics}, pages 311--318. Association for Computational
  Linguistics.

\bibitem[{Pennington et~al.(2014)Pennington, Socher, and
  Manning}]{pennington2014glove}
Jeffrey Pennington, Richard Socher, and Christopher Manning. 2014.
\newblock Glove: Global vectors for word representation.
\newblock In \emph{Proceedings of the 2014 conference on empirical methods in
  natural language processing (EMNLP)}, pages 1532--1543.

\bibitem[{Puduppully et~al.(2018)Puduppully, Dong, and
  Lapata}]{puduppully2018data}
Ratish Puduppully, Li~Dong, and Mirella Lapata. 2018.
\newblock Data-to-text generation with content selection and planning.
\newblock \emph{arXiv preprint arXiv:1809.00582}.

\bibitem[{Puzikov and Gurevych(2018)}]{puzikov2018e2e}
Yevgeniy Puzikov and Iryna Gurevych. 2018.
\newblock E2e nlg challenge: Neural models vs. templates.
\newblock In \emph{Proceedings of the 11th International Conference on Natural
  Language Generation}, pages 463--471.

\bibitem[{Reiter and Dale(2000)}]{reiter2000building}
Ehud Reiter and Robert Dale. 2000.
\newblock \emph{Building natural language generation systems}.
\newblock Cambridge university press.

\bibitem[{Schapire(1999)}]{schapire1999brief}
Robert~E Schapire. 1999.
\newblock A brief introduction to boosting.
\newblock In \emph{Ijcai}, volume~99, pages 1401--1406.

\bibitem[{Sharma et~al.(2017)Sharma, El~Asri, Schulz, and
  Zumer}]{sharma2017nlgeval}
Shikhar Sharma, Layla El~Asri, Hannes Schulz, and Jeremie Zumer. 2017.
\newblock \href {http://arxiv.org/abs/1706.09799} {Relevance of unsupervised
  metrics in task-oriented dialogue for evaluating natural language
  generation}.
\newblock \emph{CoRR}, abs/1706.09799.

\bibitem[{Shen et~al.(2017)Shen, Lei, Barzilay, and Jaakkola}]{shen2017style}
Tianxiao Shen, Tao Lei, Regina Barzilay, and Tommi Jaakkola. 2017.
\newblock Style transfer from non-parallel text by cross-alignment.
\newblock In \emph{Advances in Neural Information Processing Systems}, pages
  6830--6841.

\bibitem[{Stent et~al.(2004)Stent, Prasad, and Walker}]{stent2004trainable}
Amanda Stent, Rashmi Prasad, and Marilyn Walker. 2004.
\newblock Trainable sentence planning for complex information presentation in
  spoken dialog systems.
\newblock In \emph{Proceedings of the 42nd annual meeting on association for
  computational linguistics}, page~79. Association for Computational
  Linguistics.

\bibitem[{Vedantam et~al.(2015)Vedantam, Lawrence~Zitnick, and
  Parikh}]{vedantam2015cider}
Ramakrishna Vedantam, C~Lawrence~Zitnick, and Devi Parikh. 2015.
\newblock Cider: Consensus-based image description evaluation.
\newblock In \emph{Proceedings of the IEEE conference on computer vision and
  pattern recognition}, pages 4566--4575.

\bibitem[{Wen et~al.(2015)Wen, Gasic, Mrksic, Su, Vandyke, and
  Young}]{wen2015semantically}
Tsung-Hsien Wen, Milica Gasic, Nikola Mrksic, Pei-Hao Su, David Vandyke, and
  Steve Young. 2015.
\newblock Semantically conditioned lstm-based natural language generation for
  spoken dialogue systems.
\newblock \emph{arXiv preprint arXiv:1508.01745}.

\bibitem[{Wiseman et~al.(2017)Wiseman, Shieber, and
  Rush}]{wiseman2017challenges}
Sam Wiseman, Stuart~M Shieber, and Alexander~M Rush. 2017.
\newblock Challenges in data-to-document generation.
\newblock \emph{arXiv preprint arXiv:1707.08052}.

\bibitem[{Wiseman et~al.(2018)Wiseman, Shieber, and Rush}]{wiseman2018learning}
Sam Wiseman, Stuart~M Shieber, and Alexander~M Rush. 2018.
\newblock Learning neural templates for text generation.
\newblock \emph{arXiv preprint arXiv:1808.10122}.

\bibitem[{Zhao et~al.(2017)Zhao, Kim, Zhang, Rush, and
  LeCun}]{zhao2018adversarially}
Junbo~Jake Zhao, Yoon Kim, Kelly Zhang, Alexander~M. Rush, and Yann LeCun.
  2017.
\newblock \href {http://arxiv.org/abs/1706.04223} {Adversarially regularized
  autoencoders for generating discrete structures}.
\newblock \emph{CoRR}, abs/1706.04223.

\end{thebibliography}
